\documentclass{article}

\usepackage [
]{TeXSanity}

\usepackage{ulem}

\usepackage{subfigure}
\usepackage [preprint]{neurips_2024}

\usepackage{fancyhdr}
\fancypagestyle{plain}{
    \fancyhf{}

}
\usepackage{comment}

\usepackage{float}
\usepackage[utf8]{inputenc} 
\usepackage[T1]{fontenc}    
\usepackage{amsfonts}       
\usepackage{doi}
\usepackage{amssymb}
\usepackage{amsthm}
\usepackage{graphicx}                           
\usepackage{dcolumn}  
\usepackage[version=3]{mhchem}  
\usepackage[T1]{fontenc}                    

\usepackage{epsfig}
\usepackage{epstopdf}
\usepackage{amsbsy}
\usepackage{multirow}
\usepackage{bm}
\usepackage{siunitx}
\usepackage {CJK}
\usepackage{textcomp}
\usepackage{comment}
\usepackage{nomencl}
\usepackage{parskip}   
\usepackage{tabularx}
\usepackage{caption} 
\usepackage{todonotes}
\usepackage{tcolorbox}
\usepackage{mdframed}
\usepackage{enumitem}  
\usepackage{booktabs} 

\makeatletter                                   
\renewcommand{\fnum@figure}{Figure \thefigure} 
\makeatother                                    
\usepackage[ruled,vlined]{algorithm2e}
\usepackage{tabularx} 
\usepackage{pdflscape} 
\usepackage{bbm}
\usepackage{listings}
\usepackage{xcolor}
\usepackage{fonttable}
\usepackage{afterpage}
\tcbuselibrary{breakable, skins, listings}
\usepackage{ifxetex}
\usepackage{ifluatex}
\usepackage{verbatim}
\usepackage{courier}  
\usepackage{inconsolata}  
\providecommand{\checkmark}{{\ding{51}}}
\usepackage{pifont} 
\usepackage{amsmath}
\usepackage{geometry}
\usepackage{booktabs}  
\usepackage{float}     

\title{From Data to Theory: Autonomous Large Language Model Agents for Materials Science}

\author{ {Samuel Onimpa Alfred, Veera Sundararaghavan}\thanks{Corresponding author: Prof. Sundararaghavan, Email: veeras@umich.edu, Tel: 734-615-7242} \\
	Department of Aerospace Engineering\\
	University of Michigan\\
	Ann Arbor, MI 48109 \\
	\texttt{veeras@umich.edu} \\
}

\lstdefinestyle{matlabOutput}{
    basicstyle=\ttfamily\small,
    backgroundcolor=\color{white},
    frame=single,
    frameround=tttt,
    breaklines=true,
    numbers=left,
    numberstyle=\tiny\color{gray},
    numbersep=5pt,
    tabsize=4,
    showspaces=false,
    showstringspaces=false,
    literate={μm}{{$\mu$m}}1 
             {µ}{{$\mu$}}1
             {σ}{{$\sigma$}}1 
             {₀}{{${}_0$}}1
             {²}{{$^2$}}1
             {√}{{$\surd$}}1
             {ky}{{$k_y$}}1
             {·}{{$\cdot$}}1
             {−}{{$-$}}1
             {✓}{{$\checkmark$}}1
             {…}{{...}}1
             {σ₀}{{$\sigma_0$}}2
             {R²}{{$R^2$}}2,
    extendedchars=true,
    inputencoding=utf8,
    keepspaces=true 
}

\begin{document}
\maketitle


\section*{\centering Abstract}

We present an autonomous large language model (LLM) agent for end-to-end, data-driven materials theory development. The model can choose an equation form, generate and run its own code, and test how well the theory matches the data without human intervention. The framework combines step-by-step reasoning with expert-supplied tools, allowing the agent to adjust its approach as needed while keeping a clear record of its decisions. For well-established materials relationships such as the Hall–Petch equation and Paris law, the agent correctly identifies the governing equation and makes reliable predictions on new datasets. For more specialized relationships, such as Kuhn’s equation for the HOMO-LUMO gap of conjugated molecules as a function of length, performance depends more strongly on the underlying model, with GPT-5 showing better recovery of the correct equation. Beyond known theories, the agent can also suggest new predictive relationships, illustrated here by a strain-dependent law for changes in the HOMO-LUMO gap. At the same time, the results show that careful validation remains essential, because the agent can still return incorrect, incomplete, or inconsistent equations even when the numerical fit appears strong. Overall, these results highlight both the promise and the current limitations of autonomous LLM agents for AI-assisted scientific modeling and discovery.

\section{Introduction}
\label{sec:Introduction}

Scientific discovery has long relied on human expertise, with foundational equations such as the Hall--Petch equation \cite{hall1951deformation} in metallurgy and the Arrhenius equation \cite{arrhenius1967reaction} in chemical kinetics emerging from close attention to experimental data. In 1889, Arrhenius examined published temperature-dependent reaction-rate data and showed that several datasets could be described by a common relation. Van 't Hoff had already recognized in 1884 that the temperature dependence of equilibrium constants suggested a related mathematical form for the temperature dependence of forward and reverse reaction rates. Arrhenius then took the next step by showing that experimental rate data could be organized by a similar expression, turning an earlier theoretical idea into a broadly useful empirical law bearing his name \cite{arrhenius1967reaction}. In a similar spirit, the present work treats LLMs as knowledge-based assistants that draw on prior scientific literature to connect observed data to plausible equations before those equations are tested quantitatively.

For decades, researchers have hoped that artificial intelligence (AI) could help speed up the full scientific workflow, from data generation and hypothesis building to equation discovery and experimental validation. Early practical progress came through machine learning, which benefited from the rapid growth of materials data through high-throughput experiments \cite{hattrick2016perspective}, databases such as the Materials Project \cite{jain2013commentary}, and broader efforts such as the Materials Genome Initiative \cite{de2019new}. These tools have been applied to many problems, including prediction of metal fatigue properties \cite{alfred2025data} and thermodynamic stability of inorganic compounds \cite{zou2025predicting}. However, most machine learning models function as black boxes: they can make accurate predictions, but they do not naturally provide equations or explanations that can be turned into theory, checked against physical principles, or used to generate new hypotheses. Symbolic regression (SR), originally developed in the context of genetic programming, offers a more interpretable alternative by searching for mathematical expressions that fit data directly \cite{koza1994genetic}. Yet this search quickly becomes difficult as equation complexity grows. More recent approaches combine deep learning with symbolic search to steer the search toward physically reasonable expressions \cite{d2022deep,mundhenk2021symbolic}, but they still explore large mathematical spaces without the benefit of broad scientific knowledge, which can limit the physical meaning of the final result \cite{udrescu2020ai}.

Large language models (LLMs) introduce a different kind of capability for materials science. Because they are trained on large collections of web text, scientific papers, textbooks, and technical documents, they can draw from a wide range of scientific knowledge rather than from a single task-specific dataset. In materials science, LLM-based agents have so far been used mainly for extracting information from the literature \cite{gupta2022matscibert,kononova2021opportunities,dagdelen2024structured,polak2024extracting,gupta2024data,ansari2024agent,ghosh2026llm}. Several recent studies have shown that this can work well at scale. Dagdelen et al.\ used fine-tuned LLMs to extract structured information across multiple materials chemistry tasks \cite{dagdelen2024structured}. Polak et al.\ reported precision and recall approaching 90\% for materials data extraction using prompt-based conversational LLMs \cite{polak2024extracting}. Gupta et al.\ extracted more than one million polymer-property records from about 680{,}000 full-text articles \cite{gupta2024data}. Ansari et al.\ introduced Eunomia, a zero-shot autonomous agent with performance comparable to fine-tuned models across several extraction benchmarks \cite{ansari2024agent}. Ghosh et al.\ developed a closed-loop multi-agent system with internal consistency checks to extract thermoelectric and structural data from roughly 10{,}000 articles \cite{ghosh2026llm}. Together, these studies show that LLM-based agents can be useful in materials science, but they focus on extracting known information from the literature. They do not address the broader challenge of carrying out an end-to-end scientific fitting workflow without human intervention.

In practice, materials scientists often combine many individual experimental and computational tools, interpret intermediate results, and revise their approach when an initial plan does not work. This process is powerful, but it can also be slow, labor-intensive, and difficult to scale. The key question is therefore not only whether LLMs can help identify equations, but whether they can take part in the larger workflow used in day-to-day scientific analysis. In this work, we develop autonomous agents that decide what step to take next, use specialized tools, interpret intermediate results, and change course when needed, all while keeping a transparent record of their decisions. To the best of our knowledge, this is the first LLM-driven autonomous agent designed specifically for end-to-end data fitting in materials science. Existing fitting workflows still require human input at one or more important stages, such as choosing the equation form, selecting initial fitting parameters, judging fit quality, or deciding whether another fitting attempt is needed. We are not aware of prior work that automates this entire process through LLM-driven autonomy.

This work makes four main contributions. First, the agent generates equations from scientific reasoning rather than choosing from a fixed library of templates, making the task a direct test of scientific knowledge. Second, the framework is designed to handle unclear or imperfect LLM outputs, which are common in scientific computing tasks. Third, the agent judges its own fitting results and decides whether re-fitting or a different strategy is needed. Fourth, the framework records the full chain of decisions, making it possible to inspect every step of the fitting process. A second major contribution is the systematic evaluation of LLM scientific knowledge in the setting of autonomous data fitting. Because we intentionally remove fallback options, successful completion depends on whether the LLM can supply the governing equation from its own prior knowledge. Each case study therefore becomes a direct test of scientific reasoning: does the agent know the right equation, can it turn that equation into working code, and can it recognize when the result is physically unreasonable? By evaluating OpenAI's Generative Pre-trained Transformer (GPT) models on materials science datasets of different complexity, this work provides a quantitative picture of the current strengths and limits of LLM-based scientific knowledge as a basis for autonomous scientific workflows.

Using this framework, we study several materials science datasets governed by foundational physical relationships. These datasets serve as case studies for testing LLM-based scientific reasoning across different levels of difficulty. The main contribution of the paper is the workflow itself, which provides a general template for any field in which governing equations can be written in closed form. Performance is evaluated in terms of success rate, types of failure, and quality of reasoning, providing a clearer view of the current capabilities and limitations of LLM-driven agents for autonomous data fitting.

The remainder of this paper is organized as follows. Section~\ref{sec:The Autonomous Scientific Agent} describes the autonomous fitting framework, including the reasoning process, available tools, and validation strategy. Section~\ref{sec:Case Studies} presents the case studies, moving from successful examples to failure cases. Section~\ref{sec:Discussion} discusses implications for LLM-driven scientific workflows, current limitations, and directions for future work. Section~\ref{sec:Conclusion} presents the conclusions.

\section{The Autonomous Scientific Agent}
\label{sec:The Autonomous Scientific Agent}

\subsection{Framework Design}
A Reasoning and Acting (ReAct) loop \cite{yao2022react} is combined with a structured tool registry to form an autonomous scientific agent capable of end-to-end empirical model fitting without human intervention. The framework is domain-agnostic: while specific materials science examples are used to illustrate its operation in subsequent sections, the architecture itself is the core contribution and is readily transferable to any empirical fitting task. 

\subsection{Framework Architecture}
\begin{figure}[htbp]
\centering
\includegraphics[width=0.8\textwidth]{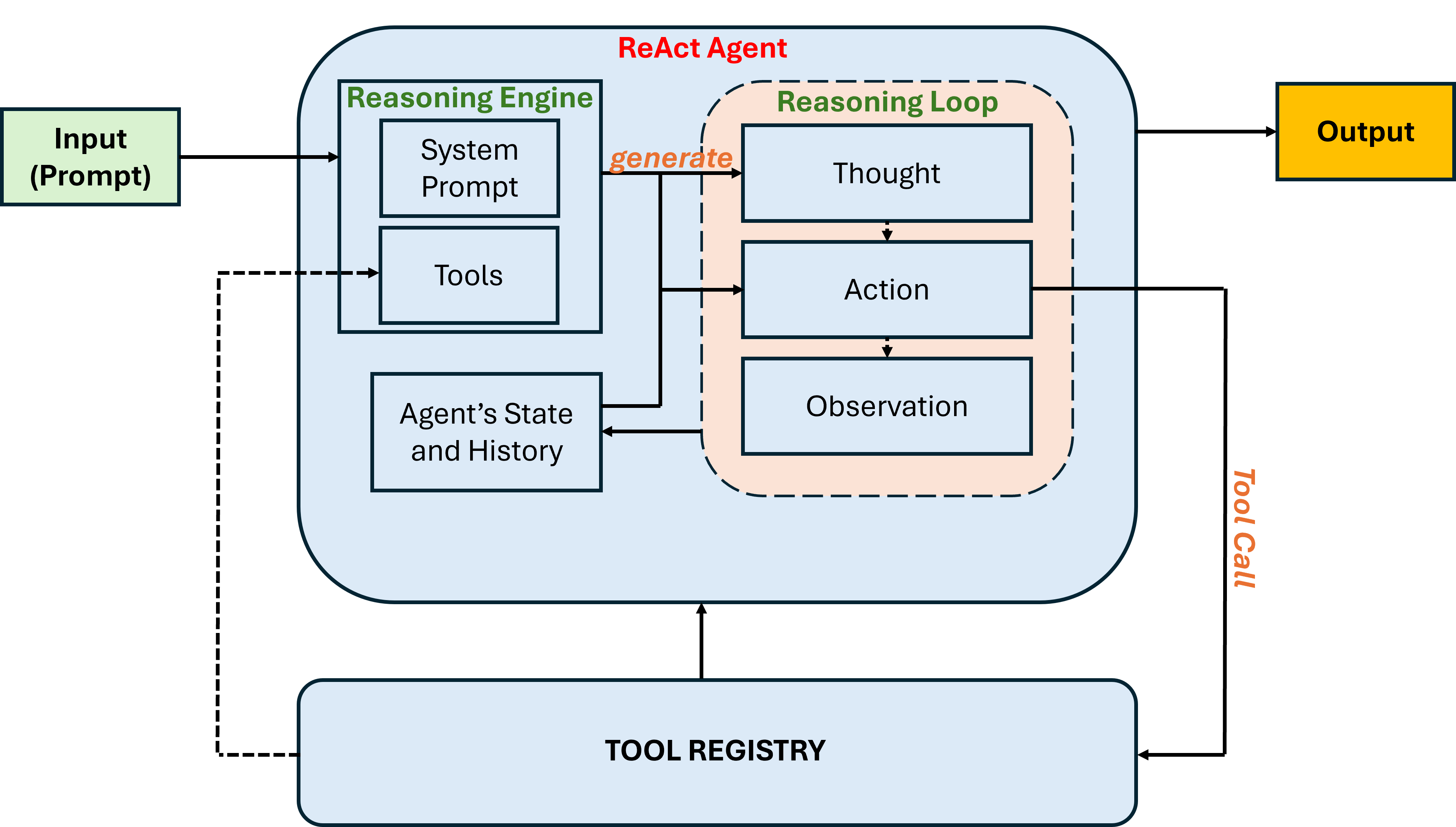}
\caption{\textit{Schematic overview of the three-component autonomous fitting framework, illustrating the closed-loop iterative workflow.} The Reasoning Engine, Tool Registry, and Agent State interact through a closed-loop iterative workflow comprising Thought, Action, and Observation steps.}
\label{fig:your_label}
\end{figure}

The framework consists of three primary components that interact iteratively to complete a fitting task, as illustrated in Fig.~\ref{fig:your_label}:
\begin{enumerate}[leftmargin=*]
    \item \textbf{Reasoning Engine:} A general-purpose LLM (e.g., GPT-4 or GPT-5) that understands task context, reasons about the current state, and selects the appropriate action to take.
    \item \textbf{Tool Registry:} A curated set of computational tools that include domain-specific operations, ranging from data loading to visualization.
    \item \textbf{Agent State:} A persistent data structure that tracks progress, stores intermediate results, and maintains a complete history of the reasoning trace.
\end{enumerate}
The agent operates within a closed-loop, iterative workflow. At each iteration, the agent executes three sequential steps: (1) \textbf{Thought}, in which the agent observes the current state and formulates a plan in natural language, resulting in a tool call; (2) \textbf{Action}, in which the agent executes the selected tool; and (3) \textbf{Observation}, in which the agent processes the tool output, which updates the state. This cycle repeats until the task is complete or a maximum iteration limit is reached.

\subsection{Large Language Models}
Two OpenAI GPT models are employed in this work: GPT-4 and GPT-5. GPT-4, a large multimodal model, demonstrated significant improvements over its predecessors in reasoning, mathematical problem-solving, and scientific knowledge retrieval \cite{achiam2023gpt}. For GPT-4, the temperature parameter is set to 0.1 to promote reproducible reasoning while allowing limited variability. GPT-5 further extends these capabilities, exhibiting enhanced reasoning, improved instruction-following, and a broader knowledge base \cite{singh2025openai}; this model is used with its default temperature settings as temperature control is not exposed in the interface. The selection of these two models enables a comparative assessment of how advances in LLM architecture and training translate to performance in scientific workflow automation, particularly in the context of symbolic function generation and empirical model fitting. 

\subsection{ReAct Reasoning Loop} 
The ReAct paradigm, originally proposed by Yao et al. \cite{yao2022react}, integrates reasoning traces with task-specific actions and is adapted here for scientific workflow automation. At each iteration, the LLM processes a structured prompt comprising four key elements: (1) the overall task description; (2) the complete tool registry with input schemas; (3) the current agent's state, including loaded data, generated functions, fitted parameters, and validation metrics; and (4) a recent history of thoughts, actions, and observations. The LLM responds in a structured format, as shown in Fig.~\ref{fig:structured_format2}:
\begin{figure}[htbp]
\centering
\begin{tcolorbox}[title=Structured Format, fonttitle=\bfseries, 
                  colback=gray!10, colframe=black]
\textcolor{Purple}{\textbf{THOUGHT:}} \textcolor{blue}{\textit{<reasoning about current state and next steps>}}\\
\textcolor{Purple}{\textbf{ACTION:}} \textcolor{blue}{\texttt{\{"tool": "<tool\_name>", "input": <JSON object>\}}}\\
\textcolor{Purple}{\textbf{OBSERVATION:}} \textcolor{blue}{\textit{<output returned by the selected tool, used to update the current agent state and inform the next iteration>}}
\end{tcolorbox}
\caption{\textit{Structured response format}. At each iteration, the LLM responds with a \textit{Thought}, an \textit{Action} specifying a tool and its input, and an \textit{Observation} that captures the tool's output.}
\label{fig:structured_format2}
\end{figure}

This structured output is parsed, validated against the tool registry, and executed. If the LLM returns malformed JSON, the agent logs the error, records the failure, and continues to the next iteration. This prevents temporary formatting mistakes from stopping the workflow. The tool's result is then added to the agent's history to inform future decisions.

\subsection{Tool Registry Design}
The tool registry provides a structured interface between the reasoning engine and the computational operations available to the agent, as illustrated in Fig.~\ref{fig:structured_format}. The reasoning engine interacts exclusively with tool descriptions and input schemas, remaining entirely isolated from the underlying implementation details. Each tool is defined by three attributes: a natural language description of its purpose, an input schema specifying the expected JSON structure for tool arguments, and an execution function implementing the corresponding computational operation. 

This modular architecture ensures extensibility, allowing new tools to be incorporated without modifying the core reasoning loop. The core set of tools common across all fitting workflows is summarized in Table~\ref{tab:tool_registry}.Additional case-specific tools can be integrated as required; for example, the fatigue crack growth case study includes an interactive region selection tool (\textit{select\_region}) to isolate the stable crack growth region of the fatigue crack growth data, while the case study on conjugated polymer band gap includes tools for extracting equations from arXiv HTML and PDF sources (\textit{extract\_equation\_html}, \textit{extract\_equation}). 

A central design principle of the proposed framework is the deliberate removal of any fallback mechanism for symbolic equation generation. In some agentic systems \cite{shinn2023reflexion,patil2023gorilla,shen2023hugginggpt}, when an LLM fails to generate a valid output, a fallback mechanism, such as a predefined template or library function, is employed to ensure task completion. While this approach improves robustness, it may obscure the true capabilities of the underlying model. The proposed framework instead enforces a strict, four-step pipeline for this critical stage, described in detail in Section ~\ref{sec:methodology}.

\begin{landscape}
\begin{figure}[htbp]
\centering
\begin{tcolorbox}[title=Sample Tool Entry, 
                  fonttitle=\bfseries,                  
                  colback=gray!10, 
                  colframe=black]

\textcolor{Purple}{\texttt{tools.generate\_function}} 
\textcolor{blue}{\texttt{= struct();}}\\
\textcolor{Purple}{\texttt{tools.generate\_function.description}} 
\textcolor{blue}{\texttt{=}}\\
\textcolor{blue}{\texttt{\hspace{1em}`Generate the Hall-Petch function using the LLM's knowledge';}}\\
\textcolor{Purple}{\texttt{tools.generate\_function.input\_schema}} 
\textcolor{blue}{\texttt{= \{\};}}\\
\textcolor{Purple}{\texttt{tools.generate\_function.execute}} 
\textcolor{Purple}{\texttt{=}}\\
\textcolor{blue}{\texttt{\hspace{1em}@(input, state) action\_generate\_function(input, state);}}

\end{tcolorbox}
\caption{\textit{Sample tool entry for the \textit{generate\_function} tool}.Each tool in the registry is defined by a description, an input schema, and an execution function.}
\label{fig:structured_format}
\end{figure}

\begin{table}[htbp]
\centering
\caption{\textit{Core tools in the agentic framework.} Each tool implements a specific operation required for empirical model fitting.}
\label{tab:tool_registry}
\begin{tabular}{llll}
\toprule
\textbf{Tool} & \textbf{Description} & \textbf{Input Schema} & \textbf{Purpose} \\
\midrule
load\_data & Load experimental data & \texttt{\{"filename": "path"\}} & Data ingestion \\
generate\_function & Generate physical equation from LLM's knowledge & \texttt{\{\}} & Symbolic function generation with no-fallback \\
fit\_model & Perform nonlinear least squares fitting & \texttt{\{"initial\_params": ...\}} & Parameter optimization \\
test\_function & Validate generated function with sample inputs & \texttt{\{"test\_params": [values], ...\}} & Prevents invalid functions \\
validate\_fit & Calculate goodness-of-fit metrics & \texttt{\{\}} & Computes R², RMSE, residuals \\
create\_plots & Generate visualizations & \texttt{\{\}} & Produces diagnostic plots \\
export\_results & Save results to file & \texttt{\{"format": "...", "filename": "..."\}} & Preserves outputs \\
finalize & Complete the task and report results & \texttt{\{\}} & Task completion \\
\bottomrule
\end{tabular}
\end{table}
\end{landscape}

\subsection{Symbolic Function Generation}
\label{sec:methodology}

The symbolic function generation pipeline is critical to the success of the framework, as it directly tests the LLM's scientific reasoning capabilities. The pipeline consist of four sequential steps:
\begin{enumerate}[leftmargin=*]
    \item \textbf{Equation Recall:}  The LLM is prompted to recall the governing equation exclusively from its parametric knowledge, without access to external sources, search tools, or retrieval-augmented generation. The agent is initialized with a task-specific system prompt that defines the agent's role, reasoning structure, and target physical context, as illustrated in Fig.~\ref{fig:prompt}.
    \item \textbf{Code Generation:}  A subsequent LLM call converts the recalled equation into executable code. The prompt specifies the required function signature, input parameters, and syntax requirements, ensuring compatibility with the computational environment.
    \item \textbf{Function Testing:}  The generated function is evaluated with sample inputs to verify syntactic correctness and basic physical plausibility, confirming that the output values are within the expected range for the given physical model.
    \item \textbf{Halt on Failure:}  If any step fails, whether it be the LLM producing an incorrect equation, generating invalid code, or producing a function that fails testing, the agent halts execution immediately. No hard-coded fallback equation is used. This strict design ensures that completed tasks reflect genuine LLM understanding, rather than a reliance on hidden fallbacks or predefined templates. The agent's reasoning trace (see Appendix~\ref{app:traces} for sample reasoning traces) provides a complete record of how the equation was generated, enabling qualitative assessment of the LLM's scientific knowledge.
\end{enumerate}

\begin{figure}[htbp]
\centering
\begin{tcolorbox}[title=LLM Initialization, 
                  fonttitle=\bfseries,                  
                  colback=gray!10,                  
                  colframe=black]

\textcolor{Purple}{\texttt{Prompt}}\texttt{ = openAIChat([}\\
\texttt{\hspace{1em}}\textcolor{blue}{\texttt{`You are an agent helping fit grain size and yield strength data}}\\
\texttt{\hspace{1em}}\textcolor{blue}{\texttt{using the Hall-Petch equation. ' ...}}\\
\texttt{\hspace{1em}}\textcolor{blue}{\texttt{`At each step: 1) Think about what to do next,}}\\
\texttt{\hspace{1em}}\textcolor{blue}{\texttt{2) Choose an action, 3) Observe results. ' ...}}\\
\texttt{\hspace{1em}}\textcolor{blue}{\texttt{`Be concise and specific.'}}\texttt{],}\\
\texttt{\hspace{1em}}\textcolor{Purple}{\texttt{ModelName}}\texttt{=}\textcolor{blue}{\texttt{`gpt-5'}}\texttt{);}

\end{tcolorbox}
\caption{\textit{Sample task-specific system prompt.} The LLM is initialized with a prompt that defines its role, the target physical context, and the expected reasoning structure.}
\label{fig:prompt}
\end{figure}

\subsection{Agent State Management}
The agent's state, which functions as working memory during execution and persistent record after completion, is organized into logical substructures as shown in Fig.~\ref{fig:agent_state}. Each tool execution updates the relevant substate and appends an entry to the history. This design serves two purposes: it provides the LLM with sufficient context for subsequent reasoning steps, and it enables post-hoc analysis of the agent's decision-making process. 
\begin{figure}[htbp]
\centering
\begin{tcolorbox}[title=Agent's State Structure, fonttitle=\bfseries, 
                  colback=gray!10, colframe=black]

\textcolor{blue}{\texttt{agent\_state.data}} \hfill \textit{\% Loaded experimental data}\\
\textcolor{blue}{\texttt{agent\_state.function}} \hfill \textit{\% Generated symbolic function and metadata}\\
\textcolor{blue}{\texttt{agent\_state.model}} \hfill \textit{\% Fitted parameters and optimization info}\\
\textcolor{blue}{\texttt{agent\_state.validation}} \hfill \textit{\% Goodness-of-fit metrics}\\
\textcolor{blue}{\texttt{agent\_state.history}} \hfill \textit{\% Complete reasoning trace}

\end{tcolorbox}
\caption{\textit{Logical substructures of the agent's state}. The state is organized into modular fields that are updated by tool executions.}
\label{fig:agent_state}
\end{figure}

\subsection{Implementation Details}
The framework is implemented in MATLAB\textsuperscript{\textregistered} R2025b \cite{MATLAB:2025b}, chosen for its robust numerical computing capabilities and native support for nonlinear optimization. GPT-4 and GPT-5 are accessed through the OpenAIChat interface within MATLAB's LLM toolbox \cite{mathworks2026llm}. Structured data exchange between tools is handled via JSON, and error handling is incorporated for malformed responses; failures are logged while allowing the agent to continue.

Tools are assigned different criticality levels. Failure of essential tools, such as \textit{generate\_function}, prompts the agent to retry before halting execution, whereas failure of auxiliary tools, such as \textit{create\_plots}, generates a warning but allows the agent to proceed. Intermediate functions generated during execution are evaluated in the base workspace, enabling seamless access by subsequent tools without re-evaluation. The complete reasoning trace is persistently stored and made available for post-hoc analysis and verification.

This domain-agnostic design fulfills the framework's primary objective: to establish a general methodology for evaluating LLM scientific reasoning across multiple domains, rather than to address any specific materials science problem.

\lstdefinestyle{traceStyle}{
  basicstyle=\ttfamily\footnotesize,
  breaklines=true,              
  breakatwhitespace=true,       
  breakindent=0pt,              
  postbreak=\mbox{$\hookrightarrow$ }, 
  columns=flexible,
  keepspaces=true,
  tabsize=2,
  numbers=none,
  frame=none,
  extendedchars=true,
  showstringspaces=false,
  xleftmargin=10pt,
  xrightmargin=10pt
}

\newtcolorbox{tracebox}{
  colback=gray!5,
  colframe=gray!30,
  rounded corners,
  arc=5pt,
  boxrule=0.5pt,
  left=5pt,
  right=5pt,
  top=10pt,
  bottom=10pt,
  breakable,
  enhanced,
  drop fuzzy shadow,
  listing engine=listings,
  listing style=traceStyle,
  listing only
}

\section{Case Studies}
\label{sec:Case Studies}

\subsection{Evaluation Framework} 
\label{sec:Evaluation Framework}
The autonomous scientific agent is evaluated on four materials science relationships: the Hall-Petch equation for grain boundary strengthening, the Paris law for fatigue crack growth, the Kuhn equation for conjugated organic chain energy gaps, and a strain-modified extension of the Kuhn equation. These case studies represent increasing levels of domain specificity and complexity, progressing from the universally known Hall-Petch relationship found in introductory textbooks to the highly specialized Kuhn equation, which appears primarily in the polymer chemistry and organic electronics literature, and finally to the strain-modified task, which lacks a canonical governing equation. This progression enables systematic assessment of LLM's scientific reasoning across varied materials science contexts.
Experimental datasets are extracted from peer-reviewed literature \cite{barnett2004influence,paul2025fatigue,sundararaghavan2023computational} for each case study, ensuring the evaluation reflects authentic materials science challenges involving real experimental scatter and non-ideal physical behavior.

Agents powered by GPT-4 and GPT-5 are evaluated across all four case studies. Performance is assessed against three criteria: whether the model correctly recalls the governing equation from parametric knowledge , without reliance on externally provided equations or hard-coded fallbacks; whether the agent successfully completes the entire fitting pipeline, from data loading through parameter estimation to statistical validation; and the nature and frequency of any failures. 

\subsection{Grain Boundary Strengthening: The Hall-Petch Relationship}

The Hall-Petch relationship, $\sigma_y = \sigma_0 + k_y d^{-1/2}$, describes grain boundary strengthening in polycrystalline materials~\cite{hall1951deformation}. This case study assesses whether the autonomous scientific agent can leverage its scientific knowledge to recall and apply the relationship to fit grain size and yield strength data extracted from Barnett et al.~\cite{barnett2004influence}. Specifically, the dataset comprises ambient temperature compression tests on Mg--3Al--1Zn magnesium alloy, with grain sizes ranging from 3 to 23~$\mu$m, showing the influence of grain size on $0.2\%$ offset yield strength.

\begin{figure}
\centering
\includegraphics[width=0.8\textwidth]{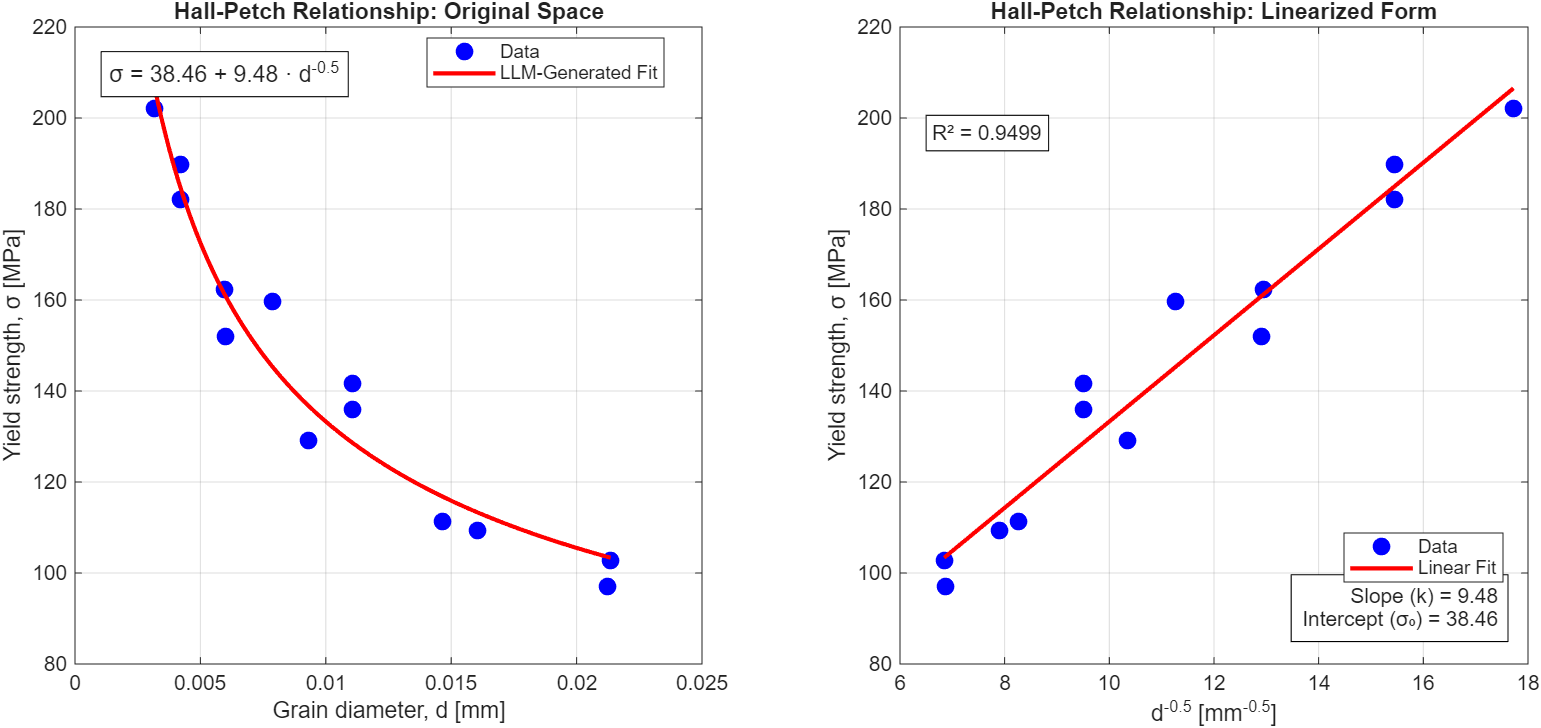}
\caption{\textit{Agent-generated Hall-Petch fitting results}.Experimental data points (blue circles) are shown together with the fitted Hall-Petch line (red line)).}
\label{fig:HP}
\end{figure}

\subsubsection{Performance Assessment}

\begin{itemize}[leftmargin=*]
\item[(a)] \textbf{Equation Recall Accuracy}: Both GPT-4 and GPT-5 generated correct symbolic forms from its parametric knowledge. GPT-4 produced $\sigma = \sigma_0 + k \cdot (1/\sqrt{d})$; GPT-5 produced the conventionally notated $\sigma = \sigma_0 + k d^{-1/2}$. Both models supplied physically accurate parameter descriptions ($\sigma_0$ as lattice friction stress, $k$ as strengthening coefficient with correct units). 

\item[(b)] \textbf{Fitting Success Rate}: Both agents completed end-to-end workflows in seven iterations, achieving statistically sound parameters. Nonlinear least squares fitting converged to identical values ($\sigma_0 = 38.46$ MPa, $k = 9.48$ MPa·$\mu$m$^{1/2}$) with $R^2 = 0.95$, metrics that indicate excellent model-data agreement. Figure~\ref{fig:HP} shows the output plot produced by the GPT-5-powered agent: experimental data points together with the fitted Hall-Petch line, demonstrating the quality of the model fit. GPT-5 exhibited marginally superior numerical intuition in initial parameter estimation ($k_0 = 10$ versus GPT-4's $k_0 = 0.05$), but this did not affect final outcomes.

\item[(c)] \textbf{Failure Mode Classification}: Both executions succeeded without failure.

\end{itemize}
\subsubsection{Insights}

The Hall-Petch case confirms that the autonomous scientific agent can execute complete fitting workflows with human-level reliability. The reasoning traces (see Appendix~\ref{app:hall_petch_traces}) reveal appropriate scientific discernment: both models correctly interpreted the inverse square root dependence, selected physically plausible initial estimates, and validated fit quality using appropriate metrics.

\subsection{Fatigue Crack Growth: The Paris Law}
\label{sec:paris_law}

The Paris law, $da/dN = C(\Delta K)^m$, describes the linear (stable) region of fatigue crack growth, where $da/dN$ is the crack growth rate per cycle, $\Delta K$ is the stress intensity factor range, and $C$ and $m$ are material constants~\cite{paris1963critical}. Unlike the previous case study, this relationship applies only to a specific subset of the fatigue crack growth dataset from Paul et al.~\cite{paul2025fatigue}, specifically the stable crack growth region (Region II). The dataset comprises FCG tests on laser powder bed fused Ti-6Al-4V specimens fabricated with a vertical orientation and tested at a stress ratio of $R = 0.1$. Consequently, the agent must first identify the appropriate data region before fitting, introducing two additional complexities: isolating the relevant data region and recognizing that the power-law relationship holds only over a limited range of $\Delta K$ values.

\subsubsection{Performance Assessment}

\begin{figure}[htbp]
\centering
\includegraphics[width=0.8\textwidth]{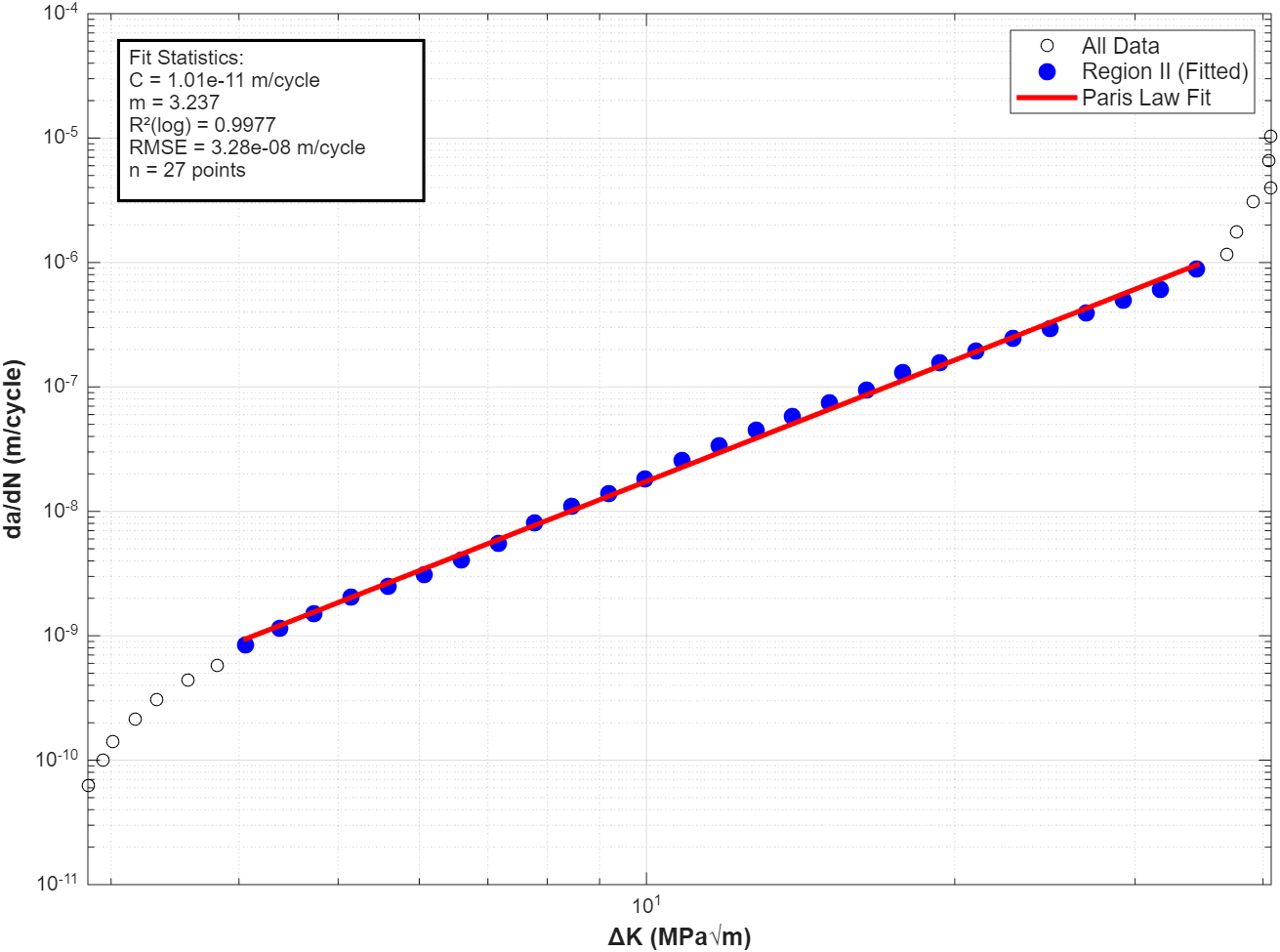}
\caption{\textit{Agent-generated Paris law fitting results.}Experimental data points (blue circles) are shown together with the fitted Paris line (red line)).}
\label{fig:PL}
\end{figure}

\begin{itemize}[leftmargin=*]
\item[(a)] \textbf{Equation Recall Accuracy}: Both GPT-4 and GPT-5 correctly generated the Paris law equation from its parametric knowledge. GPT-4 produced $da/dN = C(\Delta K)^m$ with detailed parameter descriptions and explicit unit analysis, noting that $C$ has units making the equation dimensionally consistent. GPT-5 generated an identical functional form but provided more structured parameter definitions, explicitly listing $C$ as the ``Paris law coefficient (material constant)'' and $m$ as the ``Paris law exponent (material constant)'' with comprehensive unit explanations. Both models passed the function test, confirming syntactic correctness.

\item[(b)] \textbf{Fitting Success Rate}: Both agents completed end-to-end workflows in nine iterations using automatic region selection. The \texttt{auto\_select\_region} tool identified nearly identical Region II ranges: GPT-4 selected a $\Delta K$ range of [3.808, 36.838] MPa$\sqrt{\text{m}}$ (29 points), with a detected slope range of [2.66, 3.78]; GPT-5 selected an identical $\Delta K$ range and returned the same slope detection. Initial estimates were consistent ($C \approx 9.48 \times 10^{-12}$, $m \approx 3.258$). Both agents converged to identical fitted parameters: $C = 8.7102 \times 10^{-12}$ m/cycle and $m = 3.2583$, with excellent fit quality ($R^2(\log) = 0.9963$, $R^2(\text{linear}) = 0.9949$). The fitted exponent $m = 3.2583$ falls within the expected range. Figure~\ref{fig:PL} shows the output plot generated by the GPT-5--powered agent.

\item[(c)] \textbf{Failure Mode Classification}: Both executions succeeded without failure.
\end{itemize}

\subsubsection{Insights}

The Paris law case study demonstrates that the autonomous scientific agent can handle domain-specific complexities beyond simple equation recall. Both agents correctly recognized the need to isolate Region II before fitting, a critical step distinguishing the Paris law from relationships applicable across entire data ranges. The reasoning traces (see Appendix~\ref{app:Paris_Law_traces} for the GPT-5 trace) reveal that both agents understood the physical significance of Region II selection and followed logically sequenced workflows without requiring intervention.

\subsection{Conjugated Polymer Band Gaps: The Kuhn Equation}

Kuhn's equation~\cite{kuhn1949quantum} (Eq.~\ref{eq:kuhn}) describes the relationship between the energy gap of conjugated organic systems and their chain length:
\begin{equation}
\Delta E = \frac{h^{2}}{8 m L^{2}} (N+1) + V_{0}\left(1 - \frac{1}{N}\right),
\label{eq:kuhn}
\end{equation}
where \(N = 4n+2\) is the number of \(\pi\) electrons, \(h\) is Planck's constant, \(m\) is the electron mass, \(L\) is the zig-zag length along the conjugation path, and \(V_{0}\) is the HOMO-LUMO gap of an infinitely long chain (a fitting parameter). Highly domain specific, this relationship appears primarily in the organic electronics and computational chemistry literature.

\begin{figure}[htbp]
\centering
\includegraphics[width=0.8\textwidth]{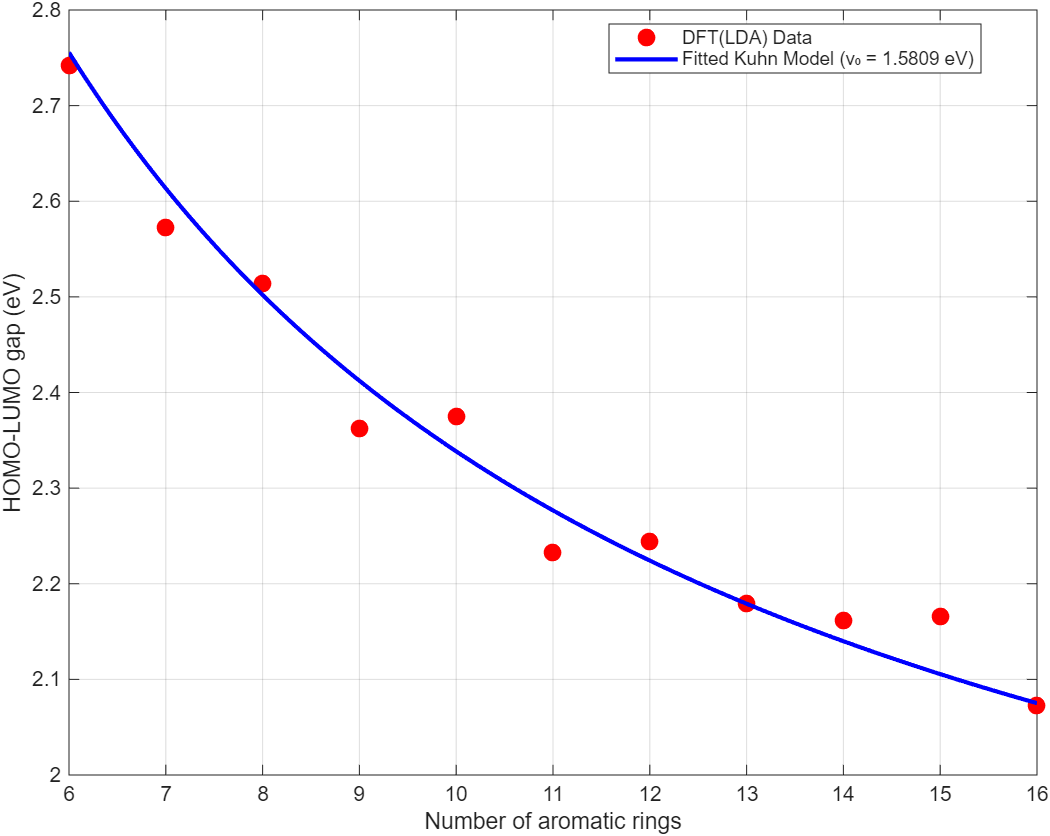}
\caption{\textit{DFT(LDA) HOMO-LUMO gaps for helicenes as a function of chain length.} The solid blue curve represents the fit provided by the GPT-5 powered agent using the Kuhn's equation.}
\label{fig:Kuhn}
\end{figure}

To evaluate the autonomous scientific agent, this case study probes two distinct capabilities: recalling the Kuhn equation from parametric knowledge and extracting and implementing it from a source document~\cite{sundararaghavan2023computational}. The source data comes from Sundararaghavan et al.~\cite{sundararaghavan2023computational}, who used density functional theory (DFT) with the local-density approximation (LDA) to compute HOMO-LUMO gaps for helicenes of varying lengths (\(n = 6\)--\(16\) benzene rings). Their results exhibit the characteristic decrease in the energy gap with increasing chain length, approaching the polymer limit (\(V_{0} \approx 1.6\) eV). Figure~\ref{fig:Kuhn} shows the DFT(LDA) data alongside the fitted curve from the GPT-5-powered agent using the generated Kuhn's equation, while Table~\ref{tab:equations} summarizes the equations generated by both agents across the knowledge recall and literature extraction tasks.

\begin{table}[htbp]
\centering
\caption{Equations generated by GPT-4 and GPT-5 powered agents across knowledge recall and literature extraction tasks.}
\label{tab:equations}
\renewcommand{\arraystretch}{2.2}  
\begin{tabular}{@{}lcc@{}}
\toprule
\textbf{Source} & \textbf{GPT-4 Powered} & \textbf{GPT-5 Powered} \\
\midrule
Knowledge recall & \(\displaystyle \frac{h^{2}}{8mL^{2}}N + V_{0}\) & \(\displaystyle \frac{h^{2}}{8mL^{2}}(N+1) + V_{0}\) \\
Literature      & \(\displaystyle \frac{h^{2}}{8mL^{2}}(N+1) + V_{0}\) & \(\displaystyle \frac{h^{2}}{8mL^{2}}(N+1) + V_{0}\left(1-\frac{1}{N}\right)\) \\
\bottomrule
\end{tabular}
\end{table}

\subsubsection{Knowledge Recall}

\textbf{(a) Equation Recall Accuracy:} When the agent relied on the underlying LLM's knowledge to recall the Kuhn equation, the GPT-4 powered agent produced \(\Delta E = \frac{h^{2}}{8mL^{2}}N + V_{0}\), which omits the \((N+1)\) factor and the entire \((1 - 1/N)\) term. The GPT-5 powered agent generated \(\Delta E = \frac{h^{2}}{8mL^{2}}(N+1) + V_{0}\), which includes the correct \((N+1)\) factor but omits the \((1 - 1/N)\) correction term. Neither agent recalled the complete canonical form from its parametric knowledge.

\textbf{(b) Fitting Success Rate:} Both agents completed end-to-end workflows in seven iterations. The GPT-4 powered agent, using its simplified equation, yielded \(V_{0} = 0.058785\) hartree (\(1.5996\) eV) with \(R^{2} = 0.9764\) and RMSE \(= 0.0303\) eV. The GPT-5 powered agent, using its partially correct equation, produced \(V_{0} = 0.058098\) hartree (\(1.5809\) eV) with \(R^{2} = 0.9712\) and RMSE \(= 0.0334\) eV. Both achieved excellent fit quality (\(R^{2} > 0.97\)) and estimated \(V_{0}\) close to the expected infinite polymer limit of \(1.6\) eV.

\textbf{(c) Failure Mode Classification:} This knowledge recall task reveals a failure mode of incomplete recall. Both agents recalled the dominant physical terms (the \(1/L^{2}\) dependence) but omitted smaller terms. The missing \((1 - 1/N)\) term becomes negligible for larger \(N\), approaching unity as chain length increases. For the dataset range of \(N = 6\)--\(16\), this term contributes at most \(16\%\) deviation for the smallest molecule and diminishes rapidly, explaining why its omission still yields excellent fit statistics. The missing \((N+1)\) factor in the GPT-4 powered agent's equation is more significant, yet its effect is partially absorbed into the fitted \(V_{0}\) parameter.

\subsubsection{Literature Extraction}

\textbf{(a) Equation Extraction Accuracy:} When required to extract the Kuhn's equation from literature, the GPT-4 powered agent extracted \(\Delta E = \frac{h^{2}}{8mL^{2}}(N+1) + V_{0}\), which still omits the \((1 - 1/N)\) term. The GPT-5 powered agent successfully extracted the complete canonical form (Eq.~\ref{eq:kuhn2}):

\begin{equation}
\Delta E = \frac{h^{2}}{8mL^{2}}(N+1) + V_{0}\left(1 - \frac{1}{N}\right)
\label{eq:kuhn2}
\end{equation}

\textbf{(b) Fitting Success Rate:} Both agents completed end-to-end extraction workflows. The GPT-4 powered agent, using its incomplete extracted equation, achieved \(V_{0} = 0.058785\) hartree (\(1.5996\) eV) with \(R^{2} = 0.9764\) and RMSE \(= 0.0303\) eV. The GPT-5 powered agent, using the complete equation, achieved \(V_{0} = 0.059506\) hartree (\(1.6193\) eV) with \(R^{2} = 0.9770\) and RMSE \(= 0.0299\) eV. Despite the missing \((1 - 1/N)\) term in the GPT-4 powered agent's equation, the fit statistics are nearly identical to those obtained with the complete form.

\textbf{(c) Failure Mode Classification:} The extraction task reveals that the \((1 - 1/N)\) correction term, while scientifically important, has a small numerical contribution over the experimental \(N\) range. For \(N = 16\) (the longest helicene in the dataset), \(1 - 1/N = 0.9375\), representing only a \(6\%\) deviation from unity. This small correction is easily absorbed into the fitted \(V_{0}\) parameter, allowing the incomplete equation to achieve statistically equivalent fits. The GPT-4 powered agent demonstrated successful extraction of the dominant terms but missed a subtle correction term, while the GPT-5 powered agent achieved complete extraction.

The incomplete equations generated by the agents can be attributed to several factors. The GPT-4 powered agent's equation \(\Delta E = \frac{h^{2}}{8mL^{2}}N + V_{0}\) represents a form of plausible hallucination, where the model generated an equation that maintains the correct physical structure but substituted \(N\) for \((N+1)\), preserving the monotonic trend while simplifying the mathematics. In the literature extraction task, the \((1 - 1/N)\) term may be presented in a way that is easily overlooked, appearing in a less prominent position or described qualitatively rather than as part of the main equation. The progression from GPT-4 to GPT-5 shows improved recall and extraction of complete scientific equations, suggesting that advances in model architecture and training data lead to better encoding of specialized domain knowledge.

\subsubsection{Insights}

The Kuhn equation case study reveals that agents can achieve excellent statistical fits even with incomplete physical models. Although theoretically important for capturing asymptotic behavior, the missing \((1 - 1/N)\) correction term contributes minimally over the experimental \(N\) range. Its omission is compensated by slight adjustments in the fitted \(V_{0}\) parameter, producing nearly identical \(R^{2}\) and RMSE values.

This finding has important implications for autonomous scientific discovery: statistical validation metrics alone cannot distinguish between complete and incomplete physical models when missing terms are numerically small. Indeed, the GPT-4-powered agent's incomplete equation, whether from knowledge recall or literature extraction, produced fit statistics indistinguishable from those of the GPT-5-powered agent's complete equation. A human scientist relying solely on goodness-of-fit measures would accept both results as equally valid.

Thus, the primary difference between the agents lies not in fit quality but in the scientific completeness of the extracted knowledge. The reliability of knowledge retrieval is fundamentally constrained by the composition and coverage of the training data: relationships that are underrepresented or require multi-step inference may be less reliably recalled~\cite{miret2024llms}. Recent studies~\cite{udrescu2020ai,shojaee2024llm,guo2025sr,song2025llm} have sought to address this by integrating LLMs within SR pipelines, leveraging domain knowledge through dimensional analysis~\cite{udrescu2020ai} and self-evaluation~\cite{ren2023self} to guide candidate equation generation. However, the assumption that LLMs possess sufficient knowledge to reliably guide equation discovery has been challenged in the literature~\cite{miret2024llms} and empirical benchmarks~\cite{shojaee2024llm}, consistent with our findings on Kuhn's equation. Shojaee et al.~\cite{shojaee2024llm} introduced LLM-SRBench, a comprehensive benchmark of 239 problems across four scientific domains, specifically designed to evaluate LLM-based equation discovery while preventing trivial memorization. Performance was assessed using symbolic accuracy, which measures recovery of the exact correct mathematical expression rather than a numerically close approximation. The best-performing system achieved only 31.5\% symbolic accuracy, revealing that LLMs struggle to generalize beyond memorized forms when confronted with genuine discovery tasks. As we show here with GPT-4 and GPT-5, as LLMs improve, one would expect corresponding improvements in knowledge retrieval. The GPT-5-powered agent's successful extraction of the full equation, including the small correction term, demonstrates a superior ability to capture subtle but physically meaningful components of scientific models. As shown in the reasoning traces (Fig.~\ref{fig:sample_trace}), the GPT-5-powered agent adaptively switched from failed PDF extraction to successful HTML extraction, while the GPT-4-powered agent continued confidently with incomplete information.

\begin{figure}[htbp]
\centering
\begin{tcolorbox}[title=Reasoning Trace, 
                  fonttitle=\bfseries,                  
                  colback=gray!10, 
                  colframe=black,
                  width=\textwidth,
                  enhanced,
                  breakable]
\lstset{basicstyle=\ttfamily\small, breaklines=true, breakatwhitespace=false}
\begin{lstlisting}
============================================================
ITERATION 3
============================================================

THOUGHT: The PDF is downloaded; next I should extract its text to locate Kuhn's equation.
ACTION: {"tool": "extract_text", "input": {}}
Executing tool: extract_text
Extracting text from: C:\Users\samoa\OneDrive - Umich\Documents\MATLAB\arXiv_2303.03490.pdf

OBSERVATION: Extracted 1 characters from PDF: C:\Users\samoa\OneDrive - Umich\Documents\MATLAB\arXiv_2303.03490.pdf

============================================================
ITERATION 4
============================================================

THOUGHT: The PDF text extraction failed (only 1 character). I should extract Kuhn's equation from the arXiv HTML instead.
ACTION: {"tool": "extract_equation_html", "input": {"paper_id": "2303.03490v1"}}
Executing tool: extract_equation_html
Successfully fetched HTML from: https://ar5iv.org/html/2303.03490
HTML truncated to 3000 characters

=== Extracted Equation ===
\Delta E = \frac{h^2}{8 m L^2}(N+1) + V_0\left(1 - \frac{1}{N}\right)
==========================

OBSERVATION: Extracted LaTeX from HTML (confidence 7/11): \Delta E = \frac{h^2}{8 m L^2}(N+1) + V_0\left(1 - \frac{1}{N}\right)
\end{lstlisting}
\end{tcolorbox}
\caption{Reasoning trace showing PDF extraction failure and subsequent successful equation extraction from HTML.}
\label{fig:sample_trace}
\end{figure}

\subsection{Strain-Modified Kuhn Equation: Helicene Response to Mechanical Deformation}

A promising role for the autonomous scientific agent is the discovery of new equations not yet reported in the literature. For example, the Kuhn equation describes the HOMO-LUMO gap of unstrained helicenes, but mechanical strain modulates this gap in an asymmetric manner: tension produces a steeper parabolic response with a maximum at 25\% strain, while compression produces a gentler inverse parabolic decrease. Unlike previous case studies, no canonical strain-modified equation exists in the literature. The agent is therefore required to generate a functional form \textit{de novo} from the LLM's knowledge, with only the base Kuhn's equation hard-coded and provided. The input prompt supplied to both GPT-4-powered and GPT-5-powered agents is shown in Fig.~\ref{fig:Kuhn_Strain_prompt}.

\begin{figure}[htbp]
\centering
\begin{tcolorbox}[title=LLM Initialization, 
                  fonttitle=\bfseries,                  
                  colback=gray!10,                  
                  colframe=black]

\textcolor{Purple}{\texttt{Prompt}}\texttt{ = openAIChat([}\\
\texttt{\hspace{1em}}\textcolor{blue}{\texttt{`You are an agent that models strain effects on helicene HOMO-LUMO gaps. ' ...}}\\
\texttt{\hspace{1em}}\textcolor{blue}{\texttt{`The base Kuhn equation is provided to you. Your task is to modify it for strain effects. ' ...}}\\
\texttt{\hspace{1em}}\textcolor{blue}{\texttt{`At each step: 1) THINK about what to do next, 2) Choose an ACTION, 3) OBSERVE the result. ' ...}}\\
\texttt{\hspace{1em}}\textcolor{blue}{\texttt{`Available tools: load\_data, generate\_strain\_function, test\_function, calculate\_response, plot\_results, finalize. ' ...}}\\
\texttt{\hspace{1em}}\textcolor{blue}{\texttt{`Respond in this EXACT format:THOUGHT: <your reasoning>ACTION: \{"tool": "<tool\_name>", "input": <json\_input>\}'}}\texttt{],}\\
\texttt{\hspace{1em}}\textcolor{Purple}{\texttt{ModelName}}\texttt{=}\textcolor{blue}{\texttt{`gpt-5'}}\texttt{);}

\end{tcolorbox}
\caption{\textit{ Prompt provided to the agent to model strain effects.} The LLM is initialized with a prompt that defines its role, the physical context (modifying the Kuhn equation for strain effects), and the expected reasoning-action-observation workflow.}
\label{fig:Kuhn_Strain_prompt}
\end{figure}

\subsubsection{Performance Assessment}

\begin{figure}[H]
\centering
\includegraphics[width=0.8\textwidth]{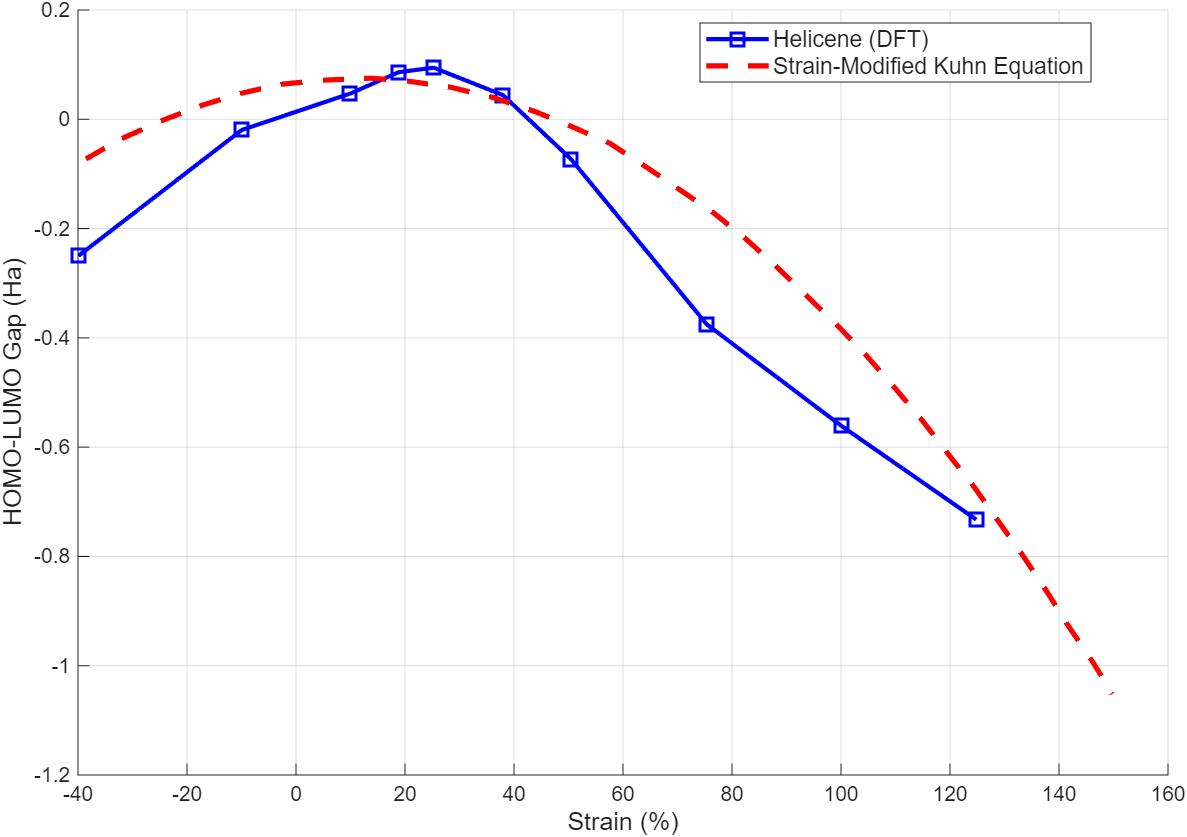}
\caption{Representative output plot from multiple agent runs showing the strain-modified Kuhn equation predictions.}
\label{fig:strain_variability}
\end{figure}

\textbf{(a) Equation Generation:} Both GPT-4-powered and GPT-5-powered agents generated a variety of functional forms across multiple runs, with no single consistent equation emerging from either agent. The GPT-4-powered agent produced variants including simple linear strain modifications and piecewise functions. The GPT-5-powered agent produced even wider variation, with functions incorporating explicit peak positioning at \(25\%\) strain, rational terms, and different coefficients for tension and compression. Both agents occasionally failed to generate any equation, and runtime was significantly longer for this case study compared to previous ones. One possible explanation for this behavior is that the agents struggle to derive a physically plausible functional form without a canonical reference, potentially resorting to hallucination or overfitting to the provided data. 
Figure~\ref{fig:strain_variability} shows a representative output plot from multiple runs. The corresponding equations produced by the GPT-5-powered agent for the strain-modified Kuhn equation (Eq.~\ref{eq:strain_kuhn}) and its associated piecewise strain function (Eq.~\ref{eq:delta}) are defined as:
\begin{equation}
\text{strain\_kuhn}(\epsilon) = \frac{h^2 (s+1)}{8m \left[ l_0 \bigl(1 + \delta(\epsilon)\bigr) \right]^2} + v_0 \bigl(1 + \gamma(\epsilon)\bigr) \left(1 - \frac{1}{s}\right)
\label{eq:strain_kuhn}
\end{equation}
where
\begin{equation}
\delta(\epsilon) =
\begin{cases}
\epsilon, & \epsilon \leq 0 \\[4pt]
\epsilon^2, & \epsilon > 0
\end{cases}
\label{eq:delta}
\end{equation}
and
\begin{equation}
\gamma(\epsilon) = 2.5\,\epsilon \left(1 - \frac{\epsilon}{0.25}\right), \quad \text{for all } \epsilon
\label{eq:gamma}
\end{equation}

\noindent Here, \(\epsilon\) is the strain, \(h\) is Planck's constant, \(m\) is the effective mass, \(s\) is the number of monomers (chain length parameter), \(l_0\) is the reference length at zero strain, and \(v_0\) is the polymer limit gap (approximately 1.6 eV). The GPT-5-powered agent fitted Eq.~\ref{eq:strain_kuhn} as shown in Fig.~\ref{fig:strain_variability} and compared it with the data from \cite{sundararaghavan2023computational}.

\textbf{(b) Workflow Efficiency:} The GPT-4-powered agent typically required more iterations (up to 10) and repeatedly called the \texttt{generate\_strain\_function} tool, showing iterative refinement. The GPT-5-powered agent completed tasks in fewer iterations (typically 4--5), generating the function in the first iteration before proceeding to data loading, calculation, and plotting.

\textbf{(c) Failure Mode Classification:} Both agents exhibited functional instability, generating different forms each run. This reflects the open-ended nature of the task: with no single correct equation, the agent explores different mathematical representations of the same physical requirements.

\subsubsection{Insights}

Both agents exhibited similar behavior: inconsistency across runs, occasional failures, and exploration of different functional spaces. This behavior reflects the inherent challenge of open-ended modeling tasks where no ground truth exists, highlighting that current LLMs may lack the consistency required for fully automated scientific workflows in such exploratory settings.

\section{Discussion}
\label{sec:Discussion}

The case studies reveal both the promise and the limitations of LLM-based autonomous agents for materials science: while they excel at executing structured workflows when scientific knowledge is reliably encoded, distinct failure modes emerge when operating at the boundaries of their parametric knowledge. This duality defines the current landscape for deploying autonomous agents in scientific workflows.

Across canonical relationships such as grain boundary strengthening and fatigue crack growth, GPT-4 and GPT-5 powered agents demonstrated reliable end-to-end fitting capabilities, correctly recalling governing equations, implementing domain-appropriate data selection strategies, and producing physically plausible parameters. This performance represents a significant advance over static pipeline approaches, suggesting that for well-established science, autonomous agents can function as capable research assistants. The Kuhn's equation case studies, however, exposed sharp reliability boundaries. When relying on knowledge recall, GPT-4 produced a scientifically incorrect yet statistically indistinguishable approximation, a plausible hallucination that passed all quantitative validation tests. When tasked with literature extraction, GPT-5 successfully adapted to retrieval failure while GPT-4 proceeded confidently with an error message. The strain modification task further revealed functional instability, with both models generating widely varying functional forms across runs, each reported with high confidence.

\subsection{Hallucination Failure Modes}

\textbf{1. Plausible hallucination.} The agent generates scientifically incorrect equations that pass standard quantitative metrics, as seen in the Kuhn's equation approximation where \(R^{2}\) values differed by only 0.0006 from the true model.

\textbf{2. Extraction failure with confident continuation.} The agent proceeds as if successful after retrieval failure, converting error messages into executable code without recognizing the underlying problem.

\textbf{3. Functional instability.} The agent generates widely varying functional forms across runs for open-ended modeling tasks, yet each run reports success.

\textbf{4. Pure fabrication.} The agent invents functional forms with no scientific basis, such as the piecewise form for the strain-modified Kuhn's equation.

A critical insight from this study is that these failure modes share a common characteristic: they produce outputs that pass conventional validation metrics while being scientifically unreliable. The near-identical fit statistics between the correct and approximate Kuhn equation formulations demonstrate that quantitative goodness-of-fit alone is insufficient to detect scientific inaccuracies. This finding raises an important question for autonomous scientific discovery: if an agent can produce scientifically incorrect results that pass standard validation metrics, how can researchers confidently trust its outputs without additional verification mechanisms?

\subsection{The Role of Self-Assessment}
The literature extraction task reveals a capability that may be as important as tool execution: recognizing failure. GPT-5 identified when PDF extraction did not extract correctly and pivoted to an alternative HTML-based strategy, successfully retrieving the correct equation. GPT-4, by contrast, treated the error message itself as valid output and proceeded to the next reasoning step. This contrast suggests that epistemic awareness, specifically the capacity to assess uncertainty, recognize knowledge boundaries, and communicate confidence levels, represents a critical direction for developing trustworthy autonomous scientific agents.
\subsection{Limitations and Future Directions}

Several opportunities for future development emerge from the findings presented in the previous sections. The current framework operates with predefined tool registries; future iterations could explore enabling agents to identify and utilize tools directly within environments such as MATLAB without explicit pre-specification. Reasoning traces, while human-readable, are not yet machine-readable, pointing to opportunities for structured output formats that support automated verification. The framework currently relies on standard fit metrics for validation; incorporating uncertainty quantification mechanisms could provide additional confidence in agent outputs. Finally, the observed variability across runs for open-ended tasks highlights the need for improved consistency to support fully autonomous operation.

These observations point toward promising research directions. Validation frameworks could extend beyond quantitative metrics to incorporate source verification, physical consistency checks, and cross-validation against independent knowledge. Hallucination detection mechanisms could target each failure mode specifically, such as testing predictions outside fitted ranges or validating retrieved content for domain-specific terminology. Epistemic awareness could be cultivated through model architecture and training, enabling agents to recognize and communicate uncertainty. Multi-agent verification offers a promising approach, where multiple agents with different underlying models cross-validate results, with discrepancies signaling targeted human review.

\section{Conclusion}
\label{sec:Conclusion}

This work presents an autonomous LLM-driven agent for end-to-end, data-driven materials theory development and fitting. Rather than serving only as a predictor, the agent can choose equation forms, generate and execute code, use specialized tools, and evaluate how well theory matches data without human intervention. The case studies show that this approach already works well for established materials relationships. For canonical laws such as the Hall--Petch equation and Paris law, both GPT-4- and GPT-5-powered agents correctly identified the governing equations, carried out the full fitting workflow, and produced reliable predictions on new datasets. For more specialized relationships, such as Kuhn's equation for the HOMO-LUMO gap of conjugated molecules, performance depended more strongly on the underlying model, with GPT-5 showing better recovery and extraction of the complete physical form. At the same time, the results make clear that strong numerical agreement does not always imply scientific correctness. In the Kuhn equation case, incomplete or incorrect equations could still produce fit statistics nearly indistinguishable from those of the correct model, showing that goodness-of-fit metrics alone are not sufficient for validation. The open-ended strain-modified Kuhn case further showed that, when no canonical equation is available, current agents can generate useful candidate relationships but still exhibit substantial variability across runs. Taken together, these findings suggest that autonomous LLM agents are already capable of supporting parts of the scientific modeling workflow, especially when the target relationship is well represented in prior knowledge. More importantly, they point toward a broader role for such systems not only in recovering known laws from data, but also in helping researchers explore and test new candidate theories. Realizing that vision, however, will require stronger verification frameworks, better handling of uncertainty, and improved consistency in open-ended discovery tasks. Although demonstrated here in materials science, the overall framework is general and can be extended to other domains in which governing relationships can be expressed in closed form. In this sense, autonomous LLM agents should be viewed not as replacements for scientific judgment, but as emerging computational partners that may help accelerate theory building, model testing, and, ultimately, the discovery of new physical laws.

\section{Acknowledgments}
The authors would like to acknowledge the Air Force Office of Scientific Research, National Science Portal Pilot program titled `Center for Scientific Machine Learning for Material Science' (Grant No. FA9550-23-1-0725) for supporting this work.

\section*{Code availability}
The code to reproduce the results is available on GitHub at \url{https://github.com/samoalfred/LLM-based-Agent}. Full reasoning traces are provided in the \href{Supplementary Material.pdf}{Supplementary Material}.

\appendix
\section{Appendix}
\label{app:traces}

\subsection{Hall-Petch: Reasoning Trace}
\label{app:hall_petch_traces}
The following output shows the complete execution trace of the autonomous agent performing Hall-Petch fitting:

\begin{lstlisting}[style=matlabOutput]


=== Autonomous Scientific Agent ===
Task: Fit Hall-Petch model to data
Data file: C:\Users\samoa\OneDrive - Umich\Documents\MATLAB\HP_Dataset.csv
Available tools: load_data, generate_function, test_function, fit_model, validate_fit, create_plots, debug_env, export_results, finalize


============================================================
ITERATION 1
============================================================

THOUGHT: The data is not loaded yet. First, load the dataset from the specified path to proceed with function generation and fitting.
ACTION: {"tool": "load_data", "input": {"filename": "C:\\Users\\samoa\\OneDrive - Umich\\Documents\\MATLAB\\HP_Dataset.csv"}}
Executing tool: load_data
Attempting to load data from: C:\Users\samoa\OneDrive - Umich\Documents\MATLAB\HP_Dataset.csv

OBSERVATION: Loaded 13 data points. d range: [0.00, 0.02] μm, σ range: [97.02, 202.05] MPa

============================================================
ITERATION 2
============================================================

THOUGHT: Data are loaded; the next required step is to generate the Hall-Petch equation using the LLM-based tool.
ACTION: {"tool": "generate_function", "input": {}}
Executing tool: generate_function
Asking LLM for Hall-Petch equation form...
✓ Received equation form:
σ = σ0 + ky d^(-1/2)

- σ: yield strength (MPa)
- σ0: grain size independent friction stress/intercept (MPa)
- ky: Hall Petch strengthening coefficient (MPa·μm^(1/2))
- d: average grain diameter (μm)
- Exponent on d: −1/2 (typical)
Converting to MATLAB function...
✓ Function test passed

OBSERVATION: LLM generated: yield_strength = @(d, sigma_0, k) sigma_0 + k .* d.^(-1/2);
Based on equation: σ = σ0 + ky d^(-1/2)

- σ: yield strength (MPa)
- σ0: grain size independent friction stress/intercept (MPa)
- ky: Hall Petch strengthening coefficient (MPa·μm^(1/2))
- d: average grain diameter (μm)
- Exponent on d: −1/2 (typical)

============================================================
ITERATION 3
============================================================

THOUGHT: Data and valid Hall-Petch function are ready and tested. Next, fit the model using reasonable initial guesses.
ACTION: {"tool": "fit_model", "input": {"initial_sigma0": 90, "initial_k": 10}}
Executing tool: fit_model

OBSERVATION: Fit complete: σ₀=38.4577, k=9.4836, resnorm=703.6743

============================================================
ITERATION 4
============================================================

THOUGHT: The model is already fitted; next, we should compute fit quality metrics before plotting and exporting.
ACTION: {"tool": "validate_fit", "input": {}}
Executing tool: validate_fit

OBSERVATION: Validation: R²=0.9499, RMSE=7.3572 MPa

============================================================
ITERATION 5
============================================================

THOUGHT: The model has been fitted and validated; next, generate visualizations (original, linearized, residuals) to assess fit quality.
ACTION: {"tool": "create_plots", "input": {}}
Executing tool: create_plots

OBSERVATION: Created 2 figures with 3 plots total. Figure 1: Original+Linearized, Figure 2: Residuals

============================================================
ITERATION 6
============================================================

THOUGHT: Fitting, validation, and plots are complete. Next, export the results for record-keeping before finalizing.
ACTION: {"tool": "export_results", "input": {"format": "table", "filename": "HP_fit_results"}}
Executing tool: export_results

OBSERVATION: Results exported to HP_fit_results

============================================================
ITERATION 7
============================================================

THOUGHT: Data loaded, function tested, model fitted and validated, plots created, and results exported. The pipeline is complete; finalize the task.
ACTION: {"tool": "finalize", "input": {}}
Executing tool: finalize

OBSERVATION: Task complete! Final model (from LLM knowledge): σ = 38.4577 + 9.4836 * d^(-0.5)
R² = 0.9499, RMSE = 7.3572 MPa
3 plots created: original space (with subplots), linearized space, and residuals
Equation source: LLM knowledge

*** Task Complete! ***

============================================================
AGENT COMPLETED
============================================================
Total iterations: 7

=== FINAL RESULTS ===
σ₀ = 38.4577 MPa
k = 9.4836 MPa·μm^0.5
R² = 0.9499
RMSE = 7.3572 MPa
Equation source: LLM knowledge
Equation: σ = 38.4577 + 9.4836 * d^(-0.5)


\end{lstlisting}

\subsection{Paris Law: Reasoning Trace}
\label{app:Paris_Law_traces}
The following output shows the complete execution trace of the autonomous agent performing Paris Law fitting:

\begin{lstlisting}[style=matlabOutput, escapeinside={(*@}{@*)}]
=== Pure ReAct Agent with Structured Tools Started (GPT-5 Optimized) ===
Task: Fit Paris law model to fatigue crack growth data
Data file: C:\Users\samoa\OneDrive - Umich\Documents\MATLAB\FCG_Data_0.1.csv
Available tools: load_data, auto_select_region, generate_function, calc_initial, fit_model, validate_fit, create_plots, debug_env, export_results, finalize

*** NOTE: Paris law equation MUST come from LLM - no fallback provided ***
*** ENHANCED PLOTTING: Will create comprehensive log-log and residual plots ***
*** AUTO-REGION SELECTION: Using slope-based detection (no manual input) ***
*** GPT-5 CONFIGURATION: Temperature=0 for deterministic output ***


============================================================
ITERATION 1
============================================================

THOUGHT: The dataset must be loaded before any region selection or fitting can occur. I will first load the CSV using the exact provided filename so subsequent tools can access the data.
ACTION: {"tool": "load_data", "input": {"filename": "C:\\Users\\samoa\\OneDrive - Umich\\Documents\\MATLAB\\FCG_Data_0.1.csv"}}
Executing tool: load_data
Attempting to load data from: C:\Users\samoa\OneDrive - Umich\Documents\MATLAB\FCG_Data_0.1.csv

OBSERVATION: Loaded 40 data points. (*@$\Delta$@*)K range: [2.848e+00, 4.074e+01] MPa(*@$\surd$@*)m, da/dN range: [6.280e-11, 1.031e-05] m/cycle

============================================================
ITERATION 2
============================================================

THOUGHT: Data are loaded but Region II is not selected yet. The next step is to automatically detect the linear Paris region before generating the Paris function and fitting.
ACTION: {"tool": "auto_select_region", "input": {}}
Executing tool: auto_select_region

OBSERVATION: Auto-selected Region II: (*@$\Delta$@*)K range [3.808, 36.838] MPa(*@$\surd$@*)m, 29 points out of 40
Detected slope range: [2.66, 3.78] (acceptable range: 2-4)

============================================================
ITERATION 3
============================================================

THOUGHT: We've loaded the data and auto-selected Region II. The next step is to generate the Paris law function from first principles: da/dN = C * ((*@$\Delta$@*)K)(*@$^m$@*), with (*@$\Delta$@*)K as the first parameter in the function signature (e.g., f((*@$\Delta$@*)K, C, m)). This will enable initial estimates and fitting.
ACTION: {"tool": "generate_function", "input": {}}
Executing tool: generate_function
Asking LLM for equation form...
(*@$\checkmark$@*) Received equation form:
1. da/dN = C((*@$\Delta$@*)K)(*@$^m$@*)
2. Parameter 1: C - Paris coefficient; an empirical, material/environment-dependent constant that sets the scale of the fatigue crack growth rate.
3. Parameter 2: m - Paris exponent; an empirical, material-dependent exponent representing the sensitivity (slope on a log-log plot) of crack growth rate to the stress-intensity-factor range.
4. Units for each variable:
- a: m
- N: cycles (dimensionless)
- da/dN: m/cycle
- (*@$\Delta$@*)K: MPa(*@$\cdot$@*)m(*@$^{0.5}$@*)
- C: (m/cycle)/(MPa(*@$\cdot$@*)m(*@$^{0.5}$@*))(*@$^m$@*)
- m: dimensionless
Converting to MATLAB function...
(*@$\checkmark$@*) Function test passed with 3 parameters
  Parameter order: dK, C, m

OBSERVATION: LLM generated: crack_growth_model = @(dK,C,m) C.*(dK).^m;
Based on equation: 1. da/dN = C((*@$\Delta$@*)K)(*@$^m$@*)
2. Parameter 1: C - Paris coefficient; an empirical, material/environment-dependent constant that sets the scale of the fatigue crack growth rate.
3. Parameter 2: m - Paris exponent; an empirical, material-dependent exponent representing the sensitivity (slope on a log-log plot) of crack growth rate to the stress-intensity-factor range.
4. Units for each variable:
- a: m
- N: cycles (dimensionless)
- da/dN: m/cycle
- (*@$\Delta$@*)K: MPa(*@$\cdot$@*)m(*@$^{0.5}$@*)
- C: (m/cycle)/(MPa(*@$\cdot$@*)m(*@$^{0.5}$@*))(*@$^m$@*)
- m: dimensionless

============================================================
ITERATION 4
============================================================

THOUGHT: Data are loaded and Region II is selected. The Paris law model function has been generated with correct parameter order. Next, we need initial estimates for C and m from the log-log linearization of Region II data to provide good starting values for nonlinear fitting.
ACTION: {"tool": "calc_initial", "input": {}}
Executing tool: calc_initial

OBSERVATION: Initial estimates: C = 9.4819e-12, m = 3.2583

============================================================
ITERATION 5
============================================================

THOUGHT: Data is loaded and Region II is selected. The Paris law function has been generated with correct parameter order ((*@$\Delta$@*)K, C, m). Initial estimates (C (*@$\approx$@*) 9.48e-12, m (*@$\approx$@*) 3.26) are within typical ranges for metals. The next logical step is to fit the model using these initial values.
ACTION: {"tool": "fit_model", "input": {"initial_C": 9.4819e-12, "initial_m": 3.2583}}
Executing tool: fit_model

OBSERVATION: Fit complete: C = 8.7102e-12, m = 3.2583, resnorm = 1.1582e-14

============================================================
ITERATION 6
============================================================

THOUGHT: The model is already fitted; the next logical step is to evaluate fit quality (R-squared, RMSE, confidence intervals) before plotting and exporting.
ACTION: {"tool": "validate_fit", "input": {}}
Executing tool: validate_fit

OBSERVATION: Validation: R(*@$^2$@*)=0.9949, R(*@$^2$@*)(log)=0.9963, RMSE (*@$\approx$@*) 1.9984e-08
Adjusted R(*@$^2$@*)=0.9946, RMSE(log) (*@$\approx$@*) 1.3744e-01

============================================================
ITERATION 7
============================================================

THOUGHT: Data are loaded, Region II is selected, the Paris law function is generated, initial parameters are estimated, the model is fitted, and metrics are validated. The next logical step is to create visualization plots (log-log fit and residuals) to assess the fit quality before exporting and finalizing.
ACTION: {"tool": "create_plots", "input": {}}
Executing tool: create_plots

OBSERVATION: Created 3 figures with comprehensive Paris law analysis. Figures 1, 2, 3

============================================================
ITERATION 8
============================================================

THOUGHT: Data are loaded, Region II selected, function generated, model fitted and validated, and plots created. The next logical step is to export the fitted parameters and metrics before finalizing.
ACTION: {"tool": "export_results", "input": {"format": "json", "filename": "FCG_Data_0.1_ParisFit.json"}}
Executing tool: export_results

OBSERVATION: Results exported to FCG_Data_0.1_ParisFit.json

============================================================
ITERATION 9
============================================================

THOUGHT: Data loaded, Region II selected, Paris function generated, initial estimates computed, model fitted, validated, plots created, and results exported. The next logical step is to finalize and present the final results.
ACTION: {"tool": "finalize", "input": {}}
Executing tool: finalize

OBSERVATION: Task complete! Final Paris law model (from LLM knowledge):
crack_growth_model = @(dK,C,m) C.*(dK).^m;
Parameter order: dK, C, m

Region automatically selected using slope analysis
Detected slope range: [2.66, 3.78]

Fitted values:
C = 8.7102e-12 m/cycle
m = 3.2583

Fit Statistics:
R(*@$^2$@*) (log) = 0.9963
R(*@$^2$@*) (linear) = 0.9949
Adjusted R(*@$^2$@*) = 0.9946
RMSE (*@$\approx$@*) 1.9984e-08 m/cycle
Region II points: 29/40
(*@$\Delta$@*)K range: [3.81, 36.84] MPa(*@$\surd$@*)m

*** Task Complete! ***

============================================================
AGENT COMPLETED
============================================================
Total iterations: 9

=== FINAL RESULTS ===
C = 8.710217e-12 m/cycle
m = 3.2583
R(*@$^2$@*) (log) = 0.9963
R(*@$^2$@*) (linear) = 0.9949
RMSE (*@$\approx$@*) 1.9984e-08 m/cycle
Equation source: LLM knowledge
Equation: crack_growth_model = @(dK,C,m) C.*(dK).^m;
Parameter order: dK, C, m
Region selection method: slope-based automatic
Detected slope range: [2.66, 3.78]
\end{lstlisting}

\bibliographystyle{unsrt}
\bibliography{refs}

\end{document}